\documentclass{article} 
\usepackage{iclr2027_conference,times}

\usepackage{amsmath,amsfonts,bm}

\def\eqref#1{equation~\ref{#1}}

\def\1{\bm{1}}

\DeclareMathAlphabet{\mathsfit}{\encodingdefault}{\sfdefault}{m}{sl}
\SetMathAlphabet{\mathsfit}{bold}{\encodingdefault}{\sfdefault}{bx}{n}

\usepackage{hyperref}
\usepackage{url}
\usepackage{booktabs}
\usepackage{multirow}
\usepackage{makecell}
\usepackage{graphicx}
\usepackage{subcaption}     
\usepackage{wrapfig}
\usepackage{amsmath, amsthm}

\newcommand{\best}[1]{\textbf{#1}}
\newcommand{\second}[1]{\underline{#1}}

\title{Adaptive Spectral-Koopman Dynamics Modeling for Temporal Domain Generalization}

\author{Tengxue Zhang$^{1}$, 
Yu Ke$^{1}$, 
Yang Shu$^{1}$, 
Chenchen Sun$^{2}$, 
Yisheng An$^{2}$, \\
\bf Chenjuan Guo$^{1}$, Bin Yang$^{1}$ \\
\small
$^{1}$East China Normal University, 
$^{2}$Chang'an University \\
}

\iclrfinalcopy 
\begin{document}

\maketitle

\begin{abstract}
Temporal Domain Generalization (TDG) has emerged to address real-world streaming data with distribution shifts over time. However, existing methods are either prone to overfitting to domain-specific noise in the data space or become overly complex and less interpretable in the parameter space. To bridge these gaps, we propose \textbf{AdaSpecK}, a spectral-Koopman framework with adaptive context extraction for TDG. To mitigate noise fitting to irregularly sampled domains, we introduce spectral-regularized Koopman dynamics modeling, which applies spectral-aware filtering in the latent space to extract denoised low-frequency trajectories and learn a Koopman operator to model the system dynamics in a linearized space. To model complex historical environments under non-stationarity, we design a context-informed heterogeneous pattern extraction mechanism. Specifically, we employ a target-conditioned attention module to attend to distinct past windows, producing a dynamic, target-specific historical summary. By constructing an environmental signature from the current evolutionary pattern, our model adaptively perceives which aspects of the past context are most informative for future prediction via a learned router. Extensive experiments on eight diverse classification and regression benchmarks demonstrate that AdaSpecK achieves state-of-the-art performance. The code and datasets are available at \href{}{https://anonymous.4open.science/r/Ada-Spec-K}.
\end{abstract}

\section{Introduction}
Conventional machine learning typically assumes that training and test data are independently and identically distributed (IID), which expects a model to generalize across training and test sets. However, this assumption is rarely satisfied in real-world applications where data often arrive as continuous streams and their underlying distributions shift dynamically over time~\citep{DBLP:conf/icml/disshift}. For instance, in financial markets, trading behaviors evolve as economic conditions change; in social media, linguistic trends and topics shift rapidly over time. This phenomenon, known as concept drift~\citep{DBLP:journals/tkde/condrift1, DBLP:conf/icml/condrift2}, leads models trained on past data to degrade when deployed in distributionally shifted environments~\citep{DBLP:journals/csur/GamaZBPB14}. This has motivated the emergence of Temporal Domain Generalization (TDG), which aims to leverage evolving historical data to build models capable of generalizing to near-future domains~\citep{DBLP:conf/nips/CaiBJ0024koodo}.

Existing approaches to TDG can be broadly categorized into data-centric and model-centric paradigms. Data-centric methods operate on the premise that temporal robustness can be achieved at the data or feature level. They employ strategies such as data augmentation~\citep{DBLP:conf/nips/manip1} or generative modeling to simulate potential future distributions~\citep{zeng2024koopmantknets}, or leverage representation learning~\citep{DBLP:conf/cvpr/rep1} to extract domain-invariant features and disentangle domain-shared features from domain-specific ones. 
In contrast, model-centric approaches focus on adapting to evolving domains by dynamically predicting or extrapolating model parameters. These methods utilize techniques such as gradient interpolation~\citep{DBLP:conf/nips/GradientI} or probabilistic frameworks to characterize latent evolutionary dynamics within the parameter space \citep{DBLP:conf/nips/CaiBJ0024koodo, yu2025frekoo}. Despite these advances, several critical challenges remain unresolved, as shown in Figure~\ref{fig:intro}:

\begin{figure}[t]
  \centering
  \includegraphics[width=0.99\linewidth]{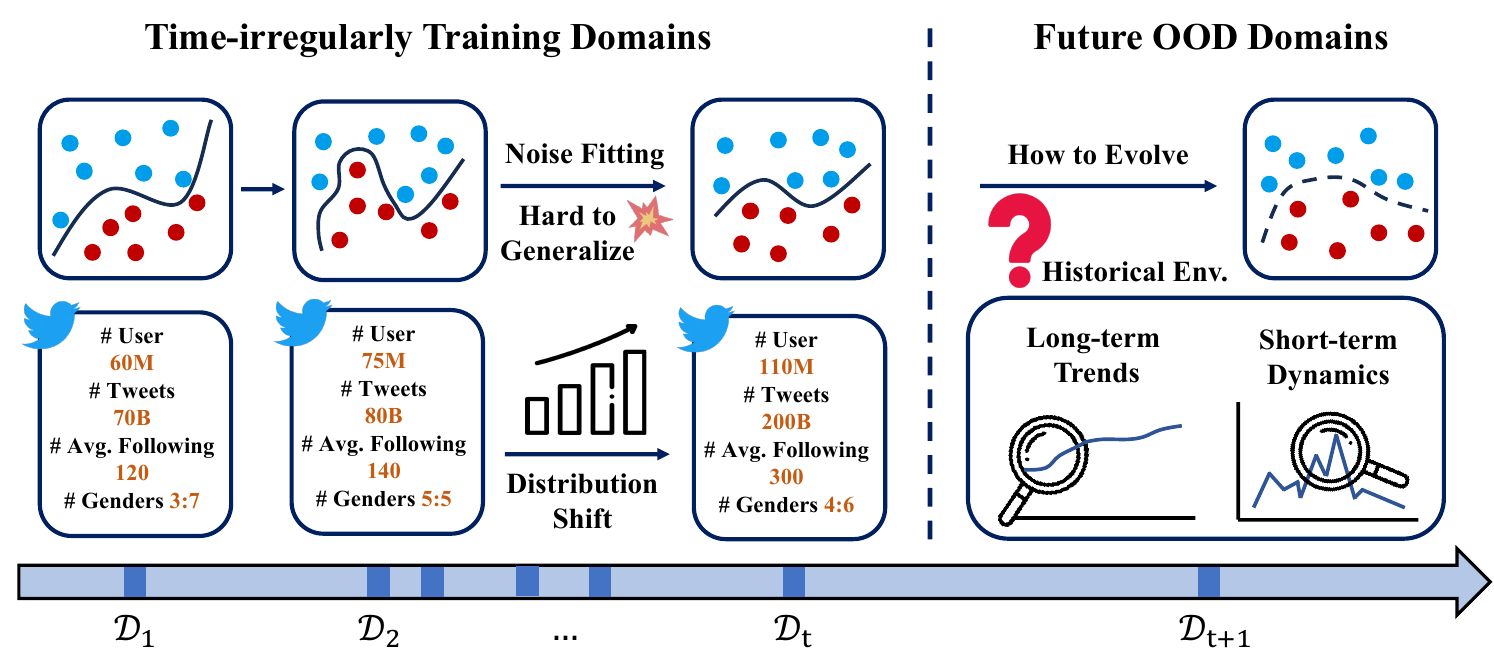}
\caption{In TDG, training domains are often time-irregularly sampled; directly learning in the data space tends to overfit domain-specific noise, impairing generalization. Predicting future OOD domains requires modeling dynamic temporal patterns across evolving historical environments.}
  \label{fig:intro}
  \vspace{-2em}
\end{figure}

\textbf{Challenge 1: Time-irregularly distributed training domains lead to noise fitting when modeling domain characteristics.} Unlike conventional forecasting tasks with regularly sampled data, TDG often involves observations at irregular timestamps. This results in training domains that are sparsely and unevenly distributed across a continuous timeline. Directly learning a predictive model in the data space risks overfitting to domain-specific patterns or noise, hindering reliable generalization to future out-of-distribution (OOD) domains~\citep{DBLP:journals/tkde/condrift1, DBLP:conf/icml/condrift2}. Although some methods instead model temporal evolution in the parameter space, this introduces its own difficulties: the parameter space is inherently more abstract and higher-dimensional, and its performance is tightly coupled to the architecture and design choices of the predictive model itself. Furthermore, the dynamics governing how parameters evolve across irregular time steps are typically highly nonlinear and lack clear structural priors, making it unstable and prone to error.

\textbf{Challenge 2: Data non-stationarity complicates the extraction of dynamic temporal patterns across historical environments.} Temporal non-stationarity, manifested as shifting data distributions and evolving feature semantics, undermines the assumption of stable statistical regularities over time~\citep{DBLP:conf/iclr/nonstat}. As a result, periodicity, seasonality, or recurring temporal structure that are readily captured under stationarity often become obfuscated or highly volatile. This renders traditional temporal modeling techniques often fail to generalize in real-world TDG scenarios \citep{DBLP:journals/csur/GamaZBPB14}. 
Moreover, future target domains may prioritize distinct contextual cues drawn from historical environments. For instance, some domains may place greater emphasis on long-term trends and macro-level patterns, while others may be highly sensitive to recent, short-term dynamics. Therefore, a robust TDG model should not only adapt to the continuous evolution of temporal patterns but also possess the flexibility to generalize across heterogeneous domain structures.

To address these challenges, we introduce \textbf{AdaSpecK}, a spectral-Koopman framework with adaptive context extraction for temporal domain generalization. 
To mitigate noise fitting to irregularly sampled training domains, we introduce spectral-regularized Koopman dynamics modeling. It applies spectral-aware filtering in the latent space to extract denoised low-frequency trajectories. By focusing on robust, slowly evolving signals, the model enhances resilience to irregular domain shifts and improves generalization to OOD scenarios. It then leverages a Koopman-style transition to model future time evolution and optimizes a Koopman operator that captures system dynamics in a linearized space. 
To address the difficulty of extracting historical information under non-stationarity, we design a context-informed heterogeneous pattern extraction mechanism. Specifically, we employ a target-conditioned attention module, which enables each future step to attend to distinct subsets of the historical windows. This effectively produces a dynamic, target-specific historical summary of past observations that adapts to the anticipated characteristics of the future domain. Furthermore, by constructing an environmental signature from the current evolutionary pattern, our model adaptively perceives which aspects of the past context are most informative for future prediction via a lightweight, learnable router. 
Overall, our main contributions are summarized as follows:

\begin{itemize}
    \item We introduce spectral-regularized Koopman dynamics modeling, which applies spectral-aware filtering in a latent space to extract denoised low-frequency trajectories and learn a Koopman operator to model the system dynamics.
      
    \item We propose a context-informed heterogeneous pattern extraction mechanism that leverages target-conditioned attention to exploit the most relevant past for each prediction step, enabling the model to selectively integrate historical contexts through adaptive expert routing.
    
    \item Extensive experiments on eight diverse benchmarks that span five classification and three regression tasks show that AdaSpecK achieves state-of-the-art performance.
\end{itemize}

\section{Related Works}

\textbf{Temporal Domain Generalization.} Domain Generalization (DG) aims to learn from multiple source domains and generalize to unseen target domains~\citep{DBLP:conf/icml/dg1, DBLP:conf/iccv/dg2, DBLP:conf/iccv/dg3, DBLP:conf/nips/dg4, DBLP:conf/nips/dg5, DBLP:conf/icdm/dg6, DBLP:conf/iclr/dg7}. DG methods can be classified into three strategies~\citep{DBLP:journals/tkde/dgcategory}: data manipulation~\citep{DBLP:conf/nips/manip1, DBLP:conf/cvpr/manip2}, representation learning~\citep{DBLP:conf/cvpr/rep1, DBLP:journals/pami/rep2}, and learning strategies such as meta-learning~\citep{DBLP:conf/cikm/meta1, DBLP:conf/aaai/meta2} and ensembles~\citep{DBLP:conf/icip/ensem1}. 
Temporal Domain Generalization (TDG) extends DG to time-evolving environments, where domain distributions shift over time. Early efforts primarily adopt data-driven strategies. 
S-MLDG~\citep{DBLP:conf/eccv/smldg} frames TDG as a sequential meta-learning problem, while TKNets~\citep{zeng2024koopmantknets} capture temporal dynamics by minimizing the divergence between predicted and actual future domain distributions for classification. Other methods focus on adapting model parameters to evolving domains. Gradient Interpolation~\citep{DBLP:conf/nips/GradientI} extrapolates model updates by supervising the first-order Taylor expansion of the learned function. LSSAE~\citep{DBLP:conf/icml/lssae} introduces a probabilistic latent space to uncover structural regularities in predictive models, and  DRAIN~\citep{bai2023drain} uses a recurrent architecture to generate time-adaptive parameters. More recently, TDG has been formalized as a continuous-time setting~\citep{DBLP:conf/nips/CaiBJ0024koodo}, with emerging work exploring frequency-domain analysis of parameter trajectories~\citep{yu2025frekoo} to better characterize temporal concept drift~\citep{DBLP:journals/csur/GamaZBPB14}. 
Despite this progress, existing methods remain limited: those operating in the data space are often prone to overfitting to domain-specific noise, while those in the parameter space tend to become overly complex and less interpretable. Moreover, none explicitly model dynamic historical environments under non-stationary conditions, overlooking how past contextual information should be adaptively leveraged for future prediction.

\textbf{Continuous Dynamics and Frequency Learning.} Continuous Dynamical Systems (CDS) model temporal evolution via differential equations, offering a natural framework for systems that evolve without discrete time steps. Neural extensions such as Neural ODEs~\citep{DBLP:conf/nips/ChenRBD18} and Latent ODEs~\citep{DBLP:journals/corr/abs-1907-03907} enable flexible, continuous-time representations in deep learning. In parallel, frequency-based learning methods analyze non-stationary temporal patterns by transforming signals into the spectral domain~\citep{DBLP:conf/nips/Autoformer, DBLP:conf/iclr/timesnet}, uncovering periodic structures and multi-scale trends that remain obscured in raw time-domain data. For instance, prior works~\citep{DBLP:conf/icml/FEDformer, DBLP:conf/nips/Fan, DBLP:conf/nips/fiternet} leverage spectral decomposition to explicitly disentangle temporal signals into low-frequency structures and high-frequency residuals or noise, and model these components separately to enhance both robustness and predictive performance.
Despite their strengths, CDS-based approaches often overlook distribution shifts and structured contextual information, while spectral methods typically lack explicit dynamic modeling. 
These limitations motivate our core insight: integrating evolving dynamics with frequency-aware representations offers a promising pathway toward improved temporal generalization.

\section{Problem Definition}
Temporal Domain Generalization (TDG) considers settings in which data distributions evolve over time, with the goal of generalizing to unseen future domains. During training, we are given a sequence of observed domains $\{D(t_1), D(t_2), \ldots, D(t_T)\}$, collected at arbitrary timestamps $T = \{t_1, t_2, \ldots, t_T\}$, where $t_i \in \mathbb{R}^+$ and $t_1 < t_2 < \ldots < t_T$. Each domain at timestamp $t$ consists of $N_t$ input-label pairs, $\mathcal{D}_t = \{ (\mathbf{x}_t^{(i)}, y_t^{(i)}) \}_{i=1}^{N_t}$, where $(\mathbf{x}_t^{(i)}, y_t^{(i)}) \in \mathcal{X}_t \times \mathcal{Y}_t$ and $\mathcal{X}_t$ and $\mathcal{Y}_t$ denote the input and label spaces at timestamp $t$. 
We assume that each domain is sampled from a time-dependent distribution $P_t(X, Y)$, and that the observed sequence undergoes \emph{distribution drift} over time. The objective is to leverage the observed domains to learn a predictive model $f(\cdot; \theta)$ that generalizes effectively to future, unseen OOD target domains associated with timestamps
$t' > t_T$.



\section{Methodology}
\begin{figure}[t]
  \centering
  \includegraphics[width=0.95\linewidth]{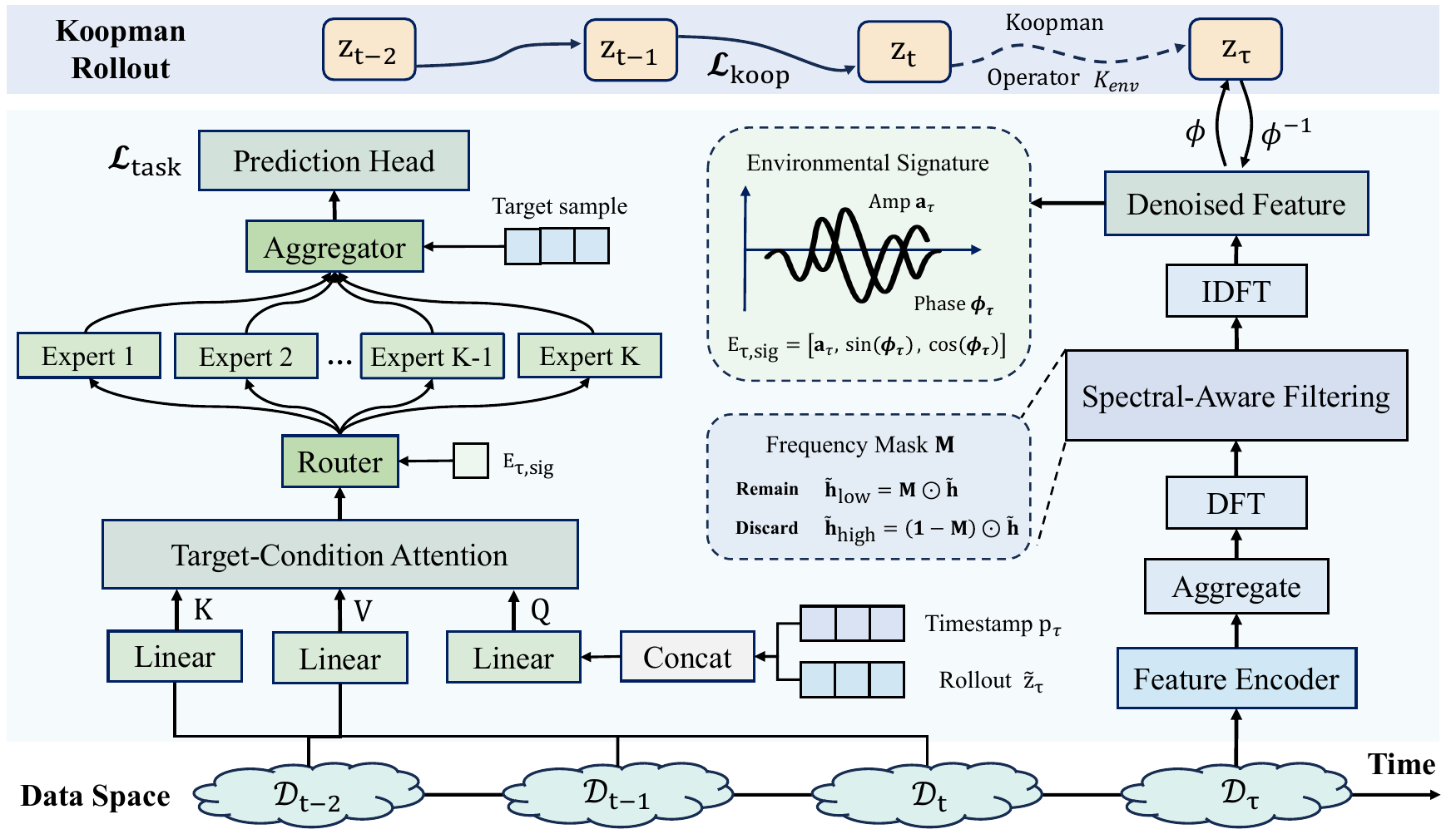}
\caption{The overview of AdaSpecK.}
  \label{fig:framework}
  \vspace{-1em}
\end{figure}

We present \textbf{AdaSpecK}, a spectral-Koopman framework with adaptive context extraction for TDG (Figure~\ref{fig:framework}). First, it employs spectral-regularized Koopman dynamics modeling to mitigate domain-specific noise fitting. By applying spectral-aware filtering in a latent space, it extracts denoised, low-frequency trajectories and learns a compact Koopman operator to model future evolution in a robust, linearized space. 
Second, it incorporates a context-informed heterogeneous pattern extraction mechanism to model complex historical environments under non-stationary. By employing a target-conditioned attention, the model generates dynamic, task-specific historical summaries that attend to distinct subsets of the historical windows. Furthermore, a learned router utilizes environmental signatures to adaptively select the most informative historical information for future prediction.

\subsection{Spectral-Regularized Koopman Dynamics Modeling}
\textbf{Spectral-Aware Filtering in Latent Space.} Direct data-space modeling in TDG often amplifies domain-specific artifacts and noise due to sparse and irregular observation timestamps.
To better capture domain structure, we first map the raw observations into a shared latent space.
For each domain $\mathcal{D}_t = \{ (\mathbf{x}_t^{(i)}, y_t^{(i)}) \}_{i=1}^{N_t}$ observed at timestamp $t$, we encode all samples within this domain using a shared feature encoder $f_\psi: \mathcal{X}_t \to \mathbb{R}^d$ and aggregate them into a domain-level representation that captures coherent domain structure while remaining robust to sample-level variability:
\begin{equation}
    h_t = \mathrm{Aggregate}\big(\{ f_\psi(\mathbf{x}_t^{(i)}) \}_{i=1}^{N_t}\big), \quad t = 1, \dots, T_c,
    \label{equ:aggre}
\end{equation}
where $\mathrm{Aggregate}(\cdot)$ denotes domain-level pooling and $T_c$ is the number of historical timestamps, yielding a latent trajectory $\mathbf{h}_{1:T_c} = [h_1, h_2, \dots, h_{T_c}]^\top \in \mathbb{R}^{T_c \times d}$.
While the latent space provides a compact representation, the trajectory remains vulnerable to transient fluctuations that may cause the model to overfit to short-term temporal irregularities. Inspired by frequency-domain insights that system-level temporal trends predominantly reside in low-frequency components \citep{DBLP:conf/icml/FEDformer,DBLP:conf/nips/Fan}, we transfer $\mathbf{h}_{1:T_c}$ to frequency domain via a Discrete Fourier Transform (DFT):
\begin{equation}
    \tilde{\mathbf{h}} = \mathcal{F}(\mathbf{h}_{1:T_c}) \in \mathbb{C}^{T_c \times d},
\end{equation}
where $\mathcal{F}(\cdot)$ denotes the DFT operator applied across the temporal dimension.
To disentangle low- and high-frequency components, we introduce a spectral ratio $\beta$ to control the proportion of preserved frequency modes. Let $k = \lfloor \beta T_c \rfloor$ denote the number of preserved frequency bins. We construct a binary mask $\mathbf{M} \in \{0,1\}^{T_c}$ that selects the $k$ lowest-magnitude frequencies:
\begin{equation}
    \mathbf{M}[i] =
    \begin{cases}
        1, & \text{if } i \in \mathcal{K}_\beta, \\
        0, & \text{otherwise},
    \end{cases}
\end{equation}
where $\mathcal{K}_\beta$ is the index set corresponding to the $k$ frequencies with the smallest magnitudes. The filtered spectrum is then obtained by applying this mask via the Hadamard product $\odot$:
\begin{align}
    \tilde{\mathbf{h}}_{\mathrm{low}} = \mathbf{M} \odot \tilde{\mathbf{h}},
     \quad\tilde{\mathbf{h}}_{\mathrm{high}} = (\mathbf{1} - \mathbf{M}) \odot \tilde{\mathbf{h}},
\end{align}
Finally, applying the inverse DFT to the low-frequency spectrum yields the denoised  trajectory:
\begin{equation}
\label{equ:inverse}
    \mathbf{h}_{1:T_c, \text{low}} = \mathcal{F}^{-1}(\tilde{\mathbf{h}}_{\mathrm{low}}) \in \mathbb{R}^{T_c \times d}.
\end{equation}
By discarding high-frequency components that are often associated with sampling noise or domain-specific perturbations, we obtain a smooth, globally coherent representation that better reflects the intrinsic temporal dynamics of the system. This spectrally filtered latent trajectory enables subsequent structured dynamics modeling to focus on robust, slowly-evolving signals, thereby enhancing resilience to irregular domain shifts and improving generalization to future OOD domains.

\textbf{Dynamics Modeling via Koopman Operator.}
Conventional temporal dynamics models, such as recurrent neural networks or models based on ordinary differential equation (ODE), often struggle with highly nonlinear systems or suffer from numerical instability during long-horizon rollouts~\citep{DBLP:conf/nips/CaiBJ0024koodo}. Koopman theory offers a powerful alternative by asserting that nonlinear dynamics can be represented as a linear system in a sufficiently high-dimensional function space~\citep{koopman1931,DBLP:journals/siamrev/koopman2022}. Building on this insight, we encode the denoised trajectory $\mathbf{h}_{1:T_c, \mathrm{low}}$ into the Koopman space via a learnable encoder $\phi$, leverage a Koopman-style transition to model their future time evolution, and decode back to the latent space via a decoder $\phi^{-1}$:
\begin{equation}
z_{t, \text{low}} = \phi(h_{t, \text{low}}),\quad 
\hat{z}_{t+1,\text{low}} = \mathbf{K}_{\mathrm{env}}z_{t, \text{low}},
\quad \hat{h}_{t+1, \text{low}} = \phi^{-1}(\hat{z}_{t+1,\text{low}}),
\label{eq:koop_roll}
\end{equation}
where $\mathbf{K}_\text{env}$ is a trainable Koopman operator that captures the intrinsic system dynamics in the Koopman space.
To ensure the Koopman operator accurately captures the underlying dynamics and maintains long-term stability, we employ a multi-faceted loss function composed of three complementary terms. First, we adopt the Koopman transition loss to enforce local fidelity of the dynamics in the Koopman space by penalizing one-step prediction errors:
\begin{equation}
\mathcal{L}_{\mathrm{koop}}
=
\sum_{t=1}^{T_c-1}
\left\| z_{t+1, \text{low}} - \hat{z}_{t+1, \text{low}} \right\|_2^2,
\end{equation}
To guarantee that the Koopman-evolved states remain grounded in the original feature manifold, we supervise the decoded rollout in the hidden space against the ground-truth hidden states:
\begin{equation}
\mathcal{L}_{\mathrm{rec}}
=
\sum_{t=1}^{T_c-1}
\left\| h_{t+1, \text{low}} - \hat{h}_{t+1, \text{low}} \right\|_2^2,
\label{eq:l_rec}
\end{equation}
To mitigate error accumulation over extended horizons, we introduce a consistency constraint that aligns the ground-truth trajectories with the dynamic evolution in the latent space:
\begin{equation}
\mathcal{L}_{\mathrm{consis}}
=
\sum_{i=1}^{T_c}\sum_{j=1}^{i}
\left\| h_{i,\text{low}} - h_{i,\text{low}}^{j \rightarrow i} \right\|_2,
\label{eq:l_consis}
\end{equation}
where $h_{i,\text{low}}^{j \rightarrow i}$ denotes the latent state at step $i$ propagated from the $j$-th anchor. This encourages the model to learn a coherent and path-independent dynamical trajectory, thereby improving robustness to temporal shifts and cumulative drift.
Overall, the three losses play different but complementary roles, encouraging the latent dynamics to be both locally accurate and globally stable.

\subsection{Context-Informed Heterogeneous Pattern Extraction}
\textbf{Target-Conditioned Historical Summary.} Data in TDG tasks often exhibit non-stationarity, as the underlying patterns evolve over time. This temporal drift complicates the modeling of dynamic temporal dependencies across shifting historical contexts and renders static, global representations inadequate for accurate long-term forecasting. To effectively capture diverse temporal dynamics and obtain robust historical patterns, we first aggregate domain-level representations $\mathbf{h}_{1:T_c}$ from past domains into a unified contextual embedding $E_\text{ctx}$:
\begin{equation}
    E_\text{ctx} = g_{\mathrm{ctx}}\big(\text{concat}[h_1, h_2, \dots,h_{T_c}]\big),
\end{equation}
where $g_{\mathrm{ctx}}$ is a lightweight multilayer perceptron (MLP).
Crucially, instead of relying on a fixed summary of the entire history, we introduce a target-conditioned attention mechanism. For each future prediction step $\tau \in \{1, \dots, H\}$, we construct a dynamic query that encapsulates its anticipated state and its absolute timestep:
\begin{equation}
    E_{\tau, q} = \text{concat}\big[ \hat{z}_{\tau},\, p_{\tau} \big],
\end{equation}
where $\hat{z}_{\tau}$ is obtained via rollout using the Koopman operator $\mathbf{K}_\text{env}$, and $p_{\tau}$ is a positional embedding that provides explicit temporal grounding. 
We then compute cross-attention between this query $E_{\tau,q}$ and the historical contextual embedding $E_\text{ctx}$:
\begin{equation}
    E_\text{history} = \text{softmax}\left(E_{\tau, q} W_Q\left(E_\text{ctx}W_K\right)^T/{\sqrt{d_k}}\right) E_\text{ctx}W_{V}
    \label{eq:cross_attn}
\end{equation}
where $ W_Q \in \mathbb{R}^{d \times d}$, $ W_K \in \mathbb{R}^{d \times d} $, $ W_V \in \mathbb{R}^{d \times d} $ are trainable projection matrices, and $d_k$ denotes the key dimension used for scaling. 
This mechanism enables the model to selectively attend to semantically and temporally relevant segments of the historical context conditioned on each future step, rather than relying on a uniform historical abstraction. Consequently, the resulting representation $E_\text{history}$ serves as a dynamic, target-specific historical summary, which better captures temporal evolution patterns and provides more precise contextual support for subsequent prediction.

\textbf{Adaptive Perception via Environmental Signature.} 
Although the historical summary provides a dynamic view of relevant past contexts, future target domains may still emphasize distinct contextual cues from historical environments. For example, some target domains may be primarily dominated by pronounced weekly cycles, whereas others may be more sensitive to recent transient events or seasonal variations. This discrepancy indicates that a universal historical encoding is insufficient for capturing heterogeneous temporal pattern extraction. Instead, the model should be able to adaptively reinterpret historical information according to the temporal characteristics of the emerging target environment.
To facilitate such adaptive perception, we first construct a compact environmental signature from the predicted future trajectory. Specifically, for each future step $\tau \in \{1, \dots, H\}$, we derive its low-frequency representation $\mathbf{h}_{\tau,\mathrm{low}}$ following Eqs.(\ref{equ:aggre})-(\ref{equ:inverse}). 

While temporal distributional shifts typically do not alter the physical mechanism of the system, they do break stable statistical regularity over time. 
As a result, periodicity, seasonality, and other temporal structures that are readily captured under stationarity may become obscured or highly volatile.
To identify relatively stable structural cues under such shifts, we extract two key features from the low-frequency components: amplitude $\mathbf{a}_{\tau}=\mathrm{Amp}(\mathbf{h}_{\tau,\mathrm{low}})$, which reflects the strength of each periodic mode, and phase $\boldsymbol{\phi}_{\tau}=\mathrm{Phase}(\mathbf{h}_{\tau,\mathrm{low}})$, which encodes temporal alignment and structural sequencing.
To ensure differentiability and avoid angular discontinuities, phase is represented via sine and cosine projections. The resulting environmental signature at time $\tau$ is defined as:
\begin{equation}
\mathrm{E}_{\tau,\mathrm{sig}}
=
\bigl[
\mathbf{a}_{\tau},\,
\sin(\boldsymbol{\phi}_{\tau}),\,
\cos(\boldsymbol{\phi}_{\tau})
\bigr].
\label{eq:sig_env}
\end{equation}
The environmental signature is then fed to an MLP-based router network $g_{\mathrm{route}}$:
\begin{equation}
\boldsymbol{w} = g_{\mathrm{route}}\bigl(\mathrm{E}_{\tau, \mathrm{sig}}\bigr),
\end{equation}
where $\boldsymbol{w} \in \mathbb{R}^K$ denotes Softmax-normalized expert weights. The target-conditioned historical summary $E_\text{history}$ is then routed via $K$ experts and aggregated into a adaptive pattern representation:
\begin{equation}
E_\text{pattern}  = \sum_{k=1}^{K} \boldsymbol{w}_{k}g_{\text{expert,k}}(E_\text{history}),
\end{equation}

To encourage experts to learn distinct and non-redundant temporal patterns, we introduce a semantic orthogonality loss $\mathcal{L}_{\mathrm{sem}}$. This loss encourages the routing weight matrix $W_{\mathrm{gate}}$ to maintain diverse semantic directions, preventing expert collapse where all experts converge to the same mean pattern:
\begin{equation}
W = W_{\mathrm{gate}}, \qquad
K = \mathrm{normalize}(W), \qquad
G = K K^{\top}, \qquad
\mathcal{L}_{\mathrm{sem}} = \mathrm{MSE}(G, I),
\label{eq:l_sem}
\end{equation}
where $W_{\mathrm{gate}}$ is the final-layer weight matrix of the router and $I$ is the identity matrix. The prediction head then fuses features of target sample $E_\text{target}$ with $E_\text{pattern}$ to yield the final prediction:
\begin{equation}
\hat{y} = g_{\mathrm{head}}([E_\text{target}, E_\text{pattern}]).
\end{equation}

Finally, the model is optimized using a composite objective comprising the task-specific loss $\mathcal{L}_{\mathrm{task}}$ with multiple regularization terms:
\begin{equation}
\mathcal{L}_{\mathrm{total}}
=
\mathcal{L}_{\mathrm{task}}
+ \lambda_{K}\bigl(\mathcal{L}_{\mathrm{koop}} + \mathcal{L}_{\mathrm{rec}}\bigr)
+ \lambda_{S}\mathcal{L}_{\mathrm{sem}}
+ \lambda_{C}\mathcal{L}_{\mathrm{consis}}.
\label{eq:total_obj}
\end{equation}
where $\mathcal{L}_{\mathrm{task}}=\sum_{i=1}^N \ell(y_i, \hat{y_i})$, using cross-entropy for classification and mean absolute error (MAE) for regression, and $\lambda_{K}$, $\lambda_{S}$, $\lambda_{C}$ are hyperparameters controlling the relative contributions.

\section{Experiments}
\subsection{Experiment Settings}
\label{main:expset}
\textbf{Datasets.}  
Following Koodos~\citep{DBLP:conf/nips/CaiBJ0024koodo} and FreKoo~\citep{yu2025frekoo}, we evaluate AdaSpecK on eight datasets spanning both classification and regression tasks. The five classification datasets include Rotated MNIST (Rot-MNIST) ~\citep{DBLP:journals/spm/mnist}, Twitter Influenza Risk (Twitter)~\citep{DBLP:journals/tkde/twitter1, DBLP:conf/sdm/twitter2}, and YearBook~\citep{yao2022wildtime}, Online News Popularity (ONP)~\citep{fernandes2015proactive}, and Shuttle~\citep{statlog_shuttle_uci}; the three regression datasets consist of Tropical Cyclone Intensity (Cyclone)~\citep{DBLP:conf/kdd/Cyclone}, House Prices (House)~\citep{decock2011ames}, and ApplianceEnergy (Appliance)~\citep{candanedo2017data}. These datasets capture diverse forms of temporal domain shift and real-world non-stationarity, providing a comprehensive benchmark for evaluation. More details of the datasets are provided in Appendix~\ref{app:data}.

\begin{table}[t]
\setlength{\tabcolsep}{1mm} 
\centering
\caption{Performance comparison on eight TDG datasets. Classification results are reported as error rates (\%) except for Twitter (AUC); regression results report MAE. ’N/A’ denotes unsupported tasks. The best and second-best results are in bold and underlined. $\uparrow$: higher is better, $\downarrow$: lower is better.}
\label{tab:main-results}
\resizebox{\linewidth}{!}{
\begin{tabular}{l|ccccc|ccc}
\toprule
\multirow{2}{*}{Model} 
& \multicolumn{5}{c|}{Classification} 
& \multicolumn{3}{c}{Regression} \\
\cmidrule(lr){2-6} \cmidrule(lr){7-9}
& Rot-MNIST $\downarrow$ & Twitter $\uparrow$ & Yearbook $\downarrow$ & ONP $\downarrow$
& Shuttle $\downarrow$ & Cyclone $\downarrow$ & House $\downarrow$ & Appliance $\downarrow$\\
\midrule

Offline 
& 6.6 $\pm$ 0.2 & 0.54 $\pm$ 0.09 & 8.6 $\pm$ 1.0 & 33.8 $\pm$ 0.6 
& 0.77 $\pm$ 0.10 & 18.7 $\pm$ 1.4 & 19.9 $\pm$ 0.1 & 10.2 $\pm$ 1.1 \\

LastDomain 
& 74.2 $\pm$ 0.9 & 0.54 $\pm$ 0.12 & 11.3 $\pm$ 1.3 & 36.0 $\pm$ 0.2 
& 0.91 $\pm$ 0.18 & 22.3 $\pm$ 0.7 & 20.6 $\pm$ 0.7 & 9.1 $\pm$ 0.7 \\

IncFinetune 
& 57.1 $\pm$ 1.4 & 0.52 $\pm$ 0.01 & 11.0 $\pm$ 0.8 & 34.0 $\pm$ 0.3 
& 0.83 $\pm$ 0.07 & 19.9 $\pm$ 0.7 & 20.6 $\pm$ 0.2 & 8.9 $\pm$ 0.5 \\

IRM 
& 8.6 $\pm$ 0.4 & 0.53 $\pm$ 0.11 & 8.3 $\pm$ 0.5 & 45.1 $\pm$ 0.1 
& 0.88 $\pm$ 0.08 & 18.0 $\pm$ 0.8 & 19.8 $\pm$ 0.2 & 5.5 $\pm$ 0.1 \\

V-REx 
& 8.6 $\pm$ 0.3 & 0.58 $\pm$ 0.05 & 8.9 $\pm$ 0.5 & 33.8 $\pm$ 0.1
& 1.90 $\pm$ 0.21 & 17.7 $\pm$ 0.5 & 20.2 $\pm$ 0.1 & 4.5 $\pm$ 0.1 \\

CIDA 
& 8.3 $\pm$ 0.7 & 0.63 $\pm$ 0.03 & 8.4 $\pm$ 0.8 & 34.7 $\pm$ 0.6 
& 7.71 $\pm$ 0.12 & 17.0 $\pm$ 0.4 & 10.2 $\pm$ 1.0 & 8.7 $\pm$ 0.2 \\

TKNets 
& 37.7 $\pm$ 2.0 & 0.57 $\pm$ 0.04 & 8.4 $\pm$ 0.3 & 33.7 $\pm$ 0.2
& 0.94 $\pm$ 0.28 & N/A & N/A & N/A \\

DRAIN 
& 59.1 $\pm$ 2.3 & 0.57 $\pm$ 0.04 & 10.5 $\pm$ 1.0 & 38.3 $\pm$ 1.2 
& 0.26 $\pm$ 0.05 & 23.6 $\pm$ 0.5 & 9.8 $\pm$ 0.1 & 6.4 $\pm$ 0.4 \\

DRAIN-$\Delta t$ 
& 57.2 $\pm$ 1.8 & 0.59 $\pm$ 0.02 & 11.0 $\pm$ 1.2 & 38.5 $\pm$ 1.1 
& 0.25 $\pm$ 0.05 & 26.2 $\pm$ 4.6 & 9.9 $\pm$ 0.1 & 6.5 $\pm$ 0.4 \\

DeepODE 
& 48.6 $\pm$ 3.2 & 0.64 $\pm$ 0.02 & 8.8 $\pm$ 2.2 & 33.6 $\pm$ 0.2
& 2.88 $\pm$ 0.71 & 18.5 $\pm$ 3.3 & 10.7 $\pm$ 0.4 & \second{4.5 $\pm$ 0.1} \\

Koodos 
& 4.8 $\pm$ 0.3
& 0.69 $\pm$ 0.02
& \second{5.9 $\pm$ 0.7} 
& \second{33.5 $\pm$ 0.4} 
& 0.24 $\pm$ 0.04
& \second{16.1 $\pm$ 1.8} 
& \second{9.0 $\pm$ 0.4} 
& 4.8 $\pm$ 0.3 \\

Frekoo 
& \second{4.7 $\pm$ 0.2} & \second{0.71 $\pm$ 0.02}  & 13.3 $\pm$ 0.8 & 34.2 $\pm$ 0.3 
& \second{0.20 $\pm$ 0.02}
& 19.5 $\pm$ 0.2 & \second{9.0 $\pm$ 0.1} & 5.4 $\pm$ 1.1 \\

\midrule
\textbf{AdaSpecK} 
& \best{4.5 $\pm$ 0.2} 
& \best{0.74 $\pm$ 0.02} 
& \best{4.5 $\pm$ 0.5} 
& \best{33.3 $\pm$ 0.2} 
& \best{0.18 $\pm$ 0.02} 
& \best{9.3 $\pm$ 0.8} 
& \best{8.9 $\pm$ 0.3} 
& \best{4.3 $\pm$ 0.1} \\

\bottomrule
\end{tabular}
}
\vspace{-1em}
\end{table}

\textbf{Baselines.}  
We compare our method against baselines from three representative categories: (1) Time-agnostic methods, including Offline~\citep{DBLP:conf/nips/CaiBJ0024koodo,yu2025frekoo}, LastDomain~\citep{DBLP:conf/nips/CaiBJ0024koodo,yu2025frekoo}, IncFinetune~\citep{DBLP:conf/nips/CaiBJ0024koodo,yu2025frekoo}, IRM~\citep{DBLP:journals/corr/abs-1907-02893}, and V-REx~\citep{DBLP:conf/icml/KruegerCJ0BZPC21}; (2) Discrete temporal domain adaptation/generalization methods, including CIDA~\citep{pmlr-v119-wang20h}, TKNets~\citep{zeng2024koopmantknets}, DRAIN, and DRAIN-$\Delta t$~\citep{bai2023drain}; and (3) Continuous temporal domain generalization methods, including DeepODE~\citep{DBLP:conf/nips/ChenRBD18,DBLP:journals/corr/abs-1907-03907}, Koodos ~\citep{DBLP:conf/nips/CaiBJ0024koodo} and Frekoo~\citep{yu2025frekoo}. For any baseline that does not support a particular dataset, we report its result as N/A in Table~\ref{tab:main-results}.

\textbf{Implementation details.}  
AdaSpecK is implemented in PyTorch and trained on an NVIDIA GeForce RTX 3090 GPU. We use the Adam optimizer with cosine annealing (learning rate of $5\times10^{-4}$, weight decay of $0.005$). We set the feature dimension $d=32$, the low-frequency ratio $\beta=0.40$. The expert count $K$ is set to 4 for Twitter, House, ONP, and Appliance, and 8 for Rotated MNIST, YearBook, Cyclone, and Shuttle. Additional details are provided in the Appendix~\ref{app:impl}.


\textbf{Metrics.} For classification tasks, we report the error rate (\%), which measures the proportion of incorrectly predicted labels. Due to severe label imbalance in the Twitter dataset, we instead use the Area Under the Curve (AUC) of the Receiver Operating Characteristic (ROC) curve. For regression tasks, performance is measured by Mean Absolute Error (MAE). All models are trained on training domains and then evaluated on unseen test domains to assess their generalization capability across temporal shifts. To ensure statistical reliability, each method is run five times with different random seeds, and we report the mean and standard deviation of the evaluation metrics across these runs.

\begin{figure}[t]
    \centering
    \begin{subfigure}[t]{0.32\linewidth}
        \centering
        \includegraphics[width=\linewidth]{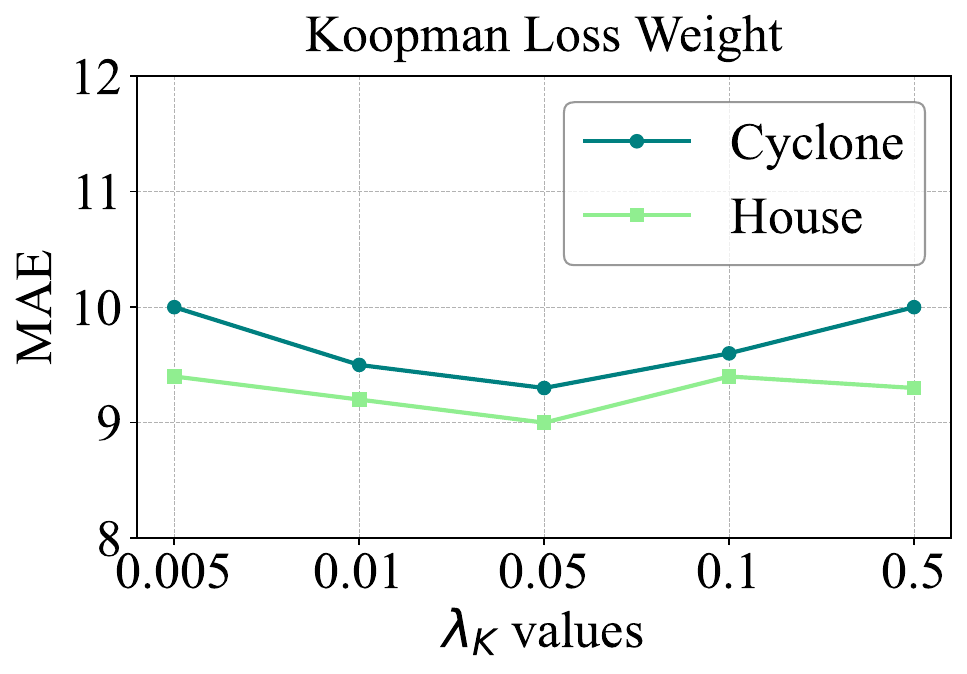}
        \caption{Sensitivity of $\lambda_{K}$}
        \label{fig:koopman loss}
    \end{subfigure}
    \hfill
    \begin{subfigure}[t]{0.32\linewidth}
        \centering
        \includegraphics[width=\linewidth]{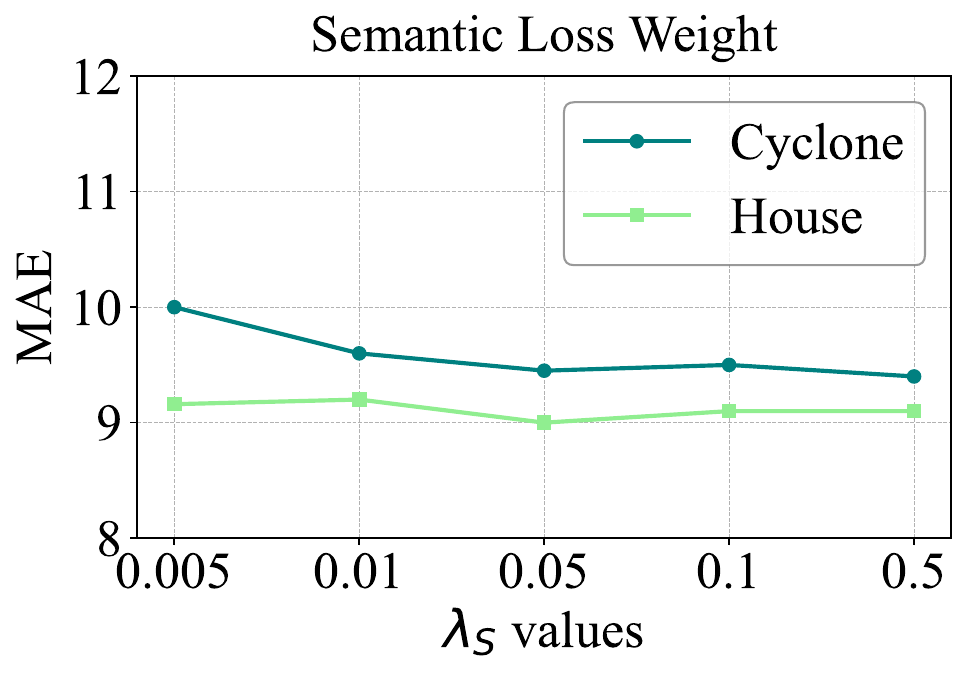}
        \caption{Sensitivity of $\lambda_{S}$}
        \label{fig:semantic loss}
    \end{subfigure}
    \hfill
    \begin{subfigure}[t]{0.32\linewidth}
        \centering
        \includegraphics[width=\linewidth]{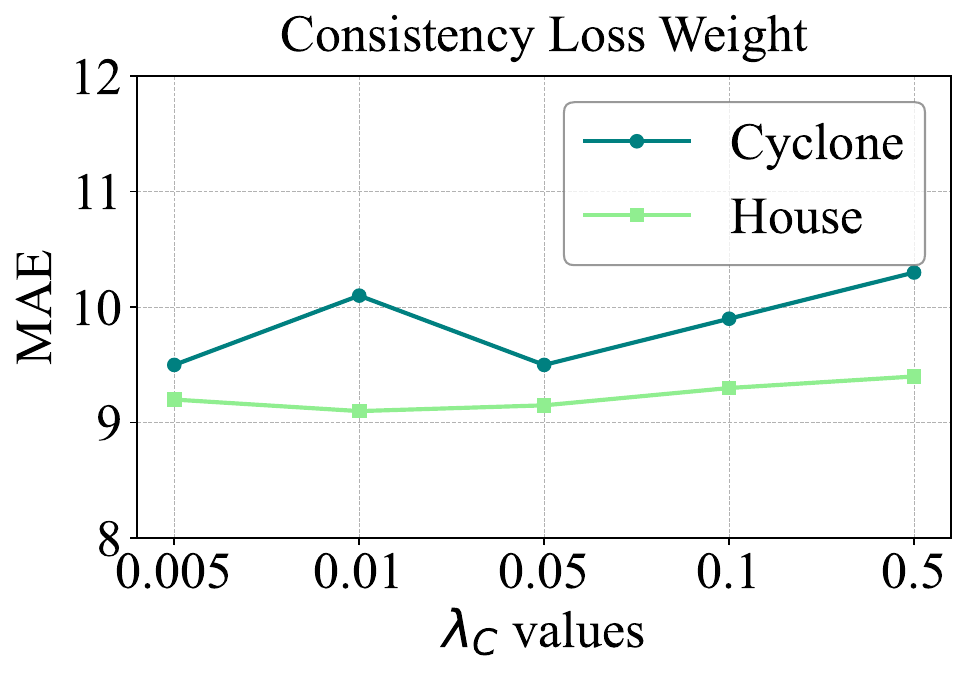}
        \caption{Sensitivity of $\lambda_{C}$}
        \label{fig:consistency loss}
    \end{subfigure}
    \vspace{-0.5em}
    \caption{
        Sensitivity analysis of loss weight $\lambda_{K}$, $\lambda_{S}$, $\lambda_{C}$.
    }
    \label{fig:loss_sensitivity}
    \vspace{-2em}
\end{figure}

\subsection{Experimental Results}
\textbf{Main Results.} Table~\ref{tab:main-results} shows that AdaSpecK achieves the best performance across all eight benchmarks, highlighting its superior capability in handling non-stationary temporal shifts. Time-agnostic baselines (e.g., Offline, LastDomain, and IncFinetune) suffer significant performance degradation on long-horizon datasets, confirming that naive reuse of historical data is inadequate under evolving temporal drift. Compared to state-of-the-art continuous TDG baselines (e.g., Koodos and Frekoo), AdaSpecK demonstrates consistent superiority. Notably, our method achieves the most substantial gains on Twitter, YearBook, Cyclone, and Appliance, where the datasets are characterized by multi-step future evolution. This resilience suggests that our learned latent dynamics provide a more stable and robust representation of temporal evolution than approaches relying solely on short-range interpolation or adjacent-domain transitions, avoiding overfitting to domain-specific artifacts or local noise. Moreover, AdaSpecK explicitly models evolving historical contexts through a context-informed, heterogeneous pattern extraction mechanism, adaptively determining how past information should be leveraged for future prediction. In summary, the proposed spectral-Koopman framework with adaptive context extraction not only improves average predictive accuracy but also enhances robustness and stability under persistent temporal distribution shifts.

\begin{wraptable}{r}{8cm}
  \centering

    \caption{Ablation study of AdaSpecK. The best and second-best results are in bold and underlined.}
      \label{tab:ablation}
    \setlength{\tabcolsep}{1mm} 
    \resizebox{\linewidth}{!}{
    \begin{tabular}{l|ccc|cc}
    \toprule
    \multirow{2}{*}{Variants}
    & \multicolumn{3}{c|}{Classification}
    & \multicolumn{2}{c}{Regression} \\
    \cmidrule(lr){2-4} \cmidrule(lr){5-6}
    & Rot-MNIST $\downarrow$ & Twitter $\uparrow$ & YearBook $\downarrow$
    & Cyclone $\downarrow$ & House $\downarrow$ \\
    \midrule
    w/o Koop.
    & 4.8 & 0.67 & \second{4.8} & 10.8 & 9.1 \\
    
    w/o Spec.
    & 4.9 & 0.63 & \second{4.8} & 10.6 & 9.0 \\
    
    w/o $E_\text{history}$
    & 4.8 & \second{0.70} & 4.9 & 10.5 & 9.2 \\
    
    w/o $\mathrm{E}_{\tau,\mathrm{sig}}$
    & \second{4.7} & 0.68 & 5.3 & \second{9.6} & \best{8.9} \\
    
    \midrule
    
    w/o $\mathcal{L}_{\mathrm{koop}}$
    & 5.1 & \second{0.69} & 6.2 & \second{9.9} & 9.0 \\
    
    w/o $\mathcal{L}_{\mathrm{rec}}$
    & \second{4.9} & 0.69 & \second{5.2} & 10.9 & 9.0 \\
    
    w/o $\mathcal{L}_{\mathrm{sem}}$
    & 5.0 & 0.65 & 5.3 & 10.5 & 9.1 \\
    
    w/o $\mathcal{L}_{\mathrm{consis}}$
    & 8.4 & 0.66 & 5.4 & 10.5 & \second{9.0} \\

    \midrule
    \textbf{AdaSpecK}
    & \best{4.5 $\pm$ 0.2}
    & \best{0.74 $\pm$ 0.02}
    & \best{4.5 $\pm$ 0.5}
    & \best{9.3 $\pm$ 0.8}
    & \best{8.9 $\pm$ 0.3} \\
    
    \bottomrule
    \end{tabular}
}
\end{wraptable}

\textbf{Ablation Study.}
Table~\ref{tab:ablation} validates the necessity of each proposed component. Regarding architecture, modeling dynamics directly in latent space without Koopman (w/o Koop.) degrades performance, confirming its role in enabling effective learning by mapping dynamics into a robust, linearized subspace. Removing spectral-aware filtering (w/o Spec.) also harms results, highlighting the importance of frequency separation for isolating slowly-varying structures critical to temporal extrapolation. Ablating dynamic target-conditioned historical summary (w/o $E_\text{history}$) or environmental signatures in the context router (w/o $\mathrm{E}_{\tau,\mathrm{sig}}$) both lead to suboptimal results, underscoring the value of adaptive historical modeling and environment-aware routing. 
Loss ablations further demonstrate the importance of each objective, with the full loss integration yielding the best results.

\subsection{Further Analysis}

\begin{figure}[t]
    \centering
    \includegraphics[width=1\linewidth]{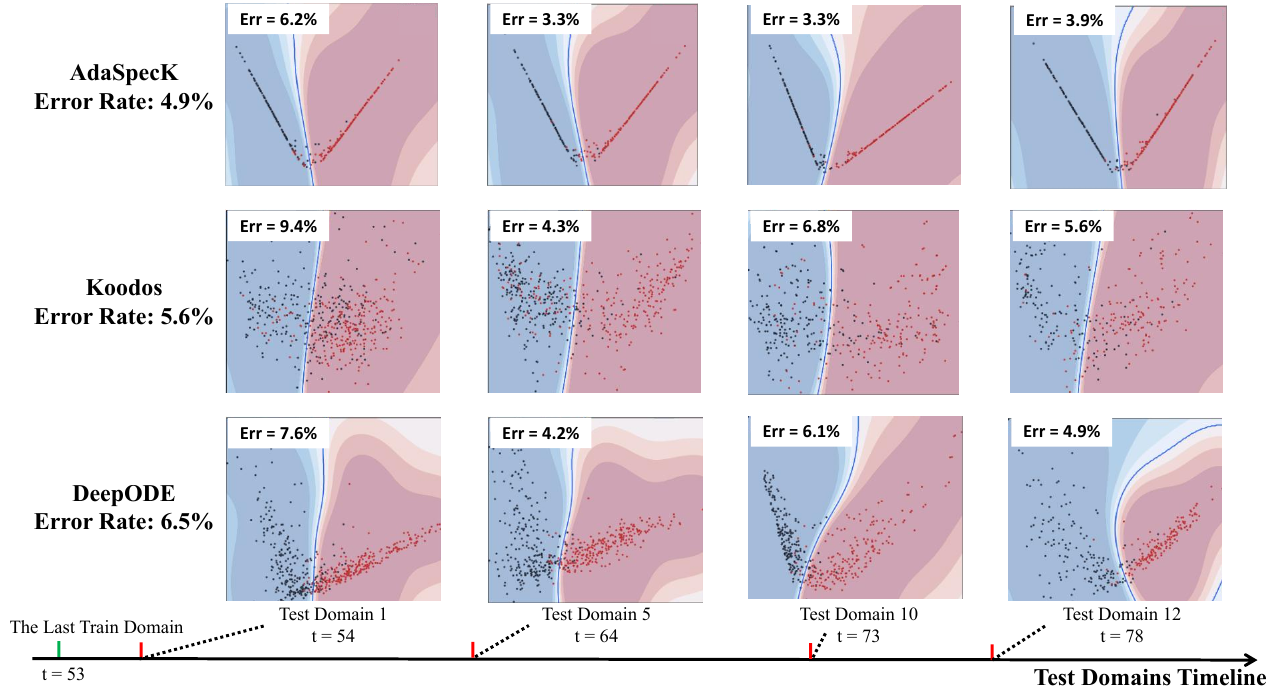}
    \vspace{-1em}
    \caption{
    Visualization of decision boundaries on YearBook. Blue and red dots represent different data classes. Rows: our method vs. baselines; columns: different test domains over time.
    }
    \label{fig:dataset_yearbook_visualization}
\end{figure} 

\textbf{Sensitivity Analysis.} 
We examine the impact of loss weights $\lambda_{K}$, $\lambda_{S}$, $\lambda_{C}$ on diverse datasets (Cyclone, House, Appliance). As shown in Figure~\ref{fig:loss_sensitivity}, a consistent pattern is that moderate coefficients lead to the best or near-best performance, whereas overly large coefficients slightly degrade accuracy, especially on Cyclone. This confirms that the auxiliary objectives serve as effective structural regularizers, with optimal results achieved under balanced weighting.



\textbf{Decision Boundary Visualization.} Figure~\ref{fig:dataset_yearbook_visualization} compares the decision regions of AdaSpecK, Koodos, and DeepODE on the YearBook test domains. 
Visually, AdaSpecK produces smoother boundaries that better reflect the underlying class separation while adapting to domain-specific shifts. In contrast, Koodos shows locally distorted or biased toward one cluster, and DeepODE fails to track evolving class geometry. The per-domain error annotations further show that AdaSpecK maintains consistent performance across target domains. 
Together, these findings demonstrate that AdaSpecK’s integration of spectral-regularized Koopman dynamics modeling and target-conditioned context extraction effectively captures temporally evolving decision structures.

\begin{figure}[h]
\centering
    \centering
    \begin{subfigure}[t]{0.325\linewidth}
        \centering
        \includegraphics[width=\linewidth]{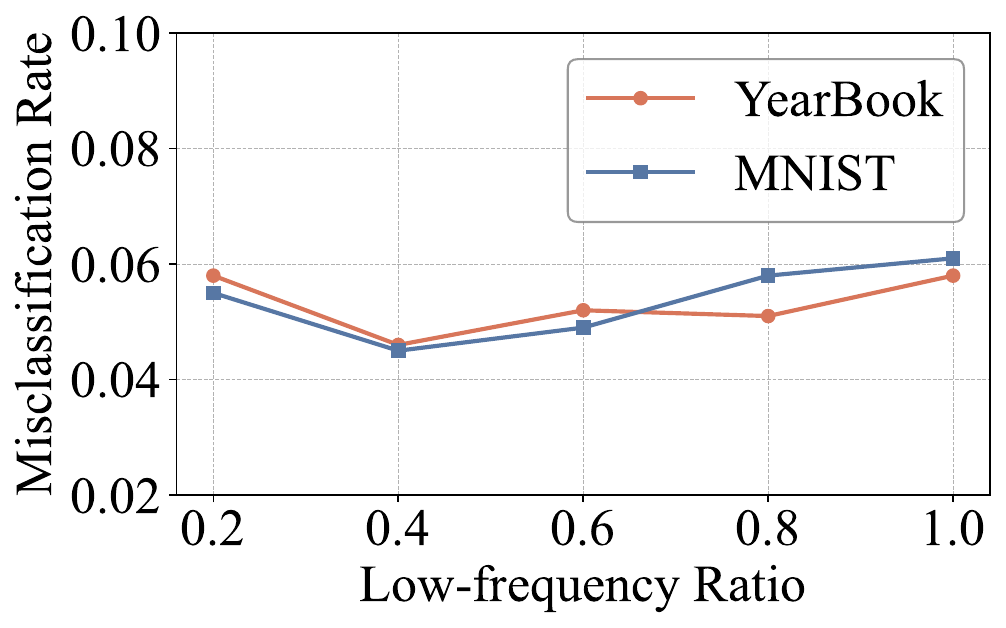}
        \caption{Impact of Spectral Ratio}
        \label{fig:spectral}
    \end{subfigure}
     \hfill
    \begin{subfigure}[t]{0.325\linewidth}
        \centering
        \includegraphics[width=\linewidth]{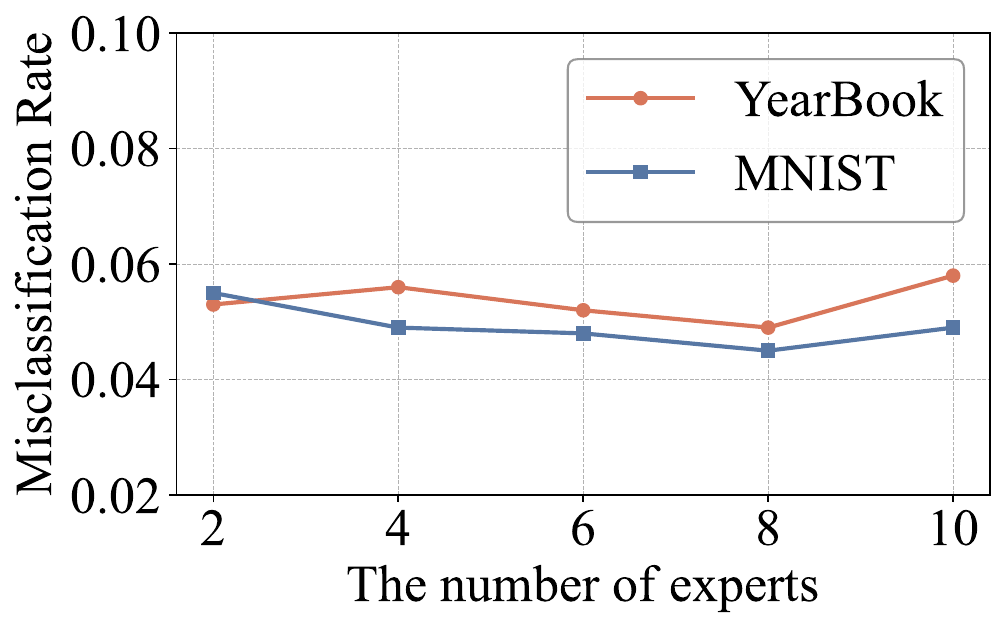}
        \caption{MNIST \& YearBook}
        \label{fig:moenum_classify}
    \end{subfigure}
    \hfill
    \begin{subfigure}[t]{0.31\linewidth}
        \centering
        \includegraphics[width=\linewidth]{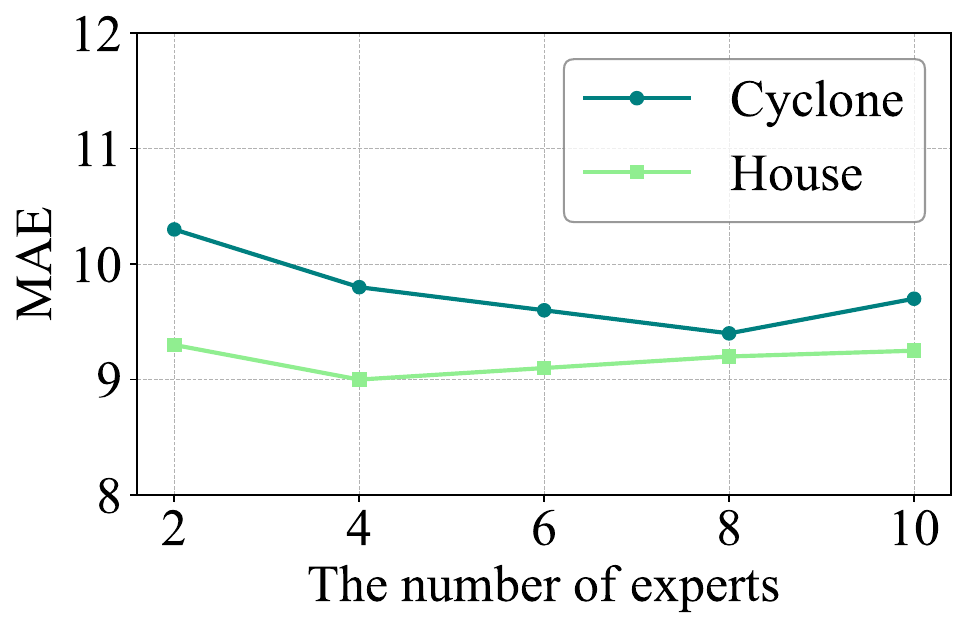}
        \caption{Cyclone \& House}
        \label{fig:moenum_regress}
    \end{subfigure}
    \vspace{-0.5em}
   \caption{(a) Impact of the spectral ratio $\beta$; (b, c) Choice of the number of experts.}
    \label{fig:MoE_number}
    \vspace{-1.5em}
\end{figure}

\textbf{Impact of Spectral ratio.} 
Figure~\ref{fig:spectral} illustrates the effect of the spectral cutoff ratio $\beta$. We observe that performance is suboptimal with an excessively small $\beta$, while a moderate ratio yields peak performance. However, further increasing $\beta$ leads to a noticeable decline in accuracy. This trend aligns with the design principles of spectral decomposition: the low-frequency branch must preserve denoised slowly-varying temporal structures conducive to Koopman dynamic modeling, while preventing the inclusion of high-frequency noise that could impair model stability.

\textbf{Choice of Expert Numbers.} 
We study the impact of expert number $K\in[2,4,6, 8, 10]$ (Figure~\ref{fig:moenum_classify} and~\ref{fig:moenum_regress}). A small $K$ fails to adequately capture the diversity of heterogeneous historical patterns, while a large $K$ increases computational cost and complexity, potentially leading to overfitting and degraded results.  Overall, $K$ in the range of 4-8 offers the best balance and is ideal for practical use.

\section{Conclusion}

This paper proposes \textbf{AdaSpecK}, a spectral-Koopman framework with adaptive context extraction for temporal domain generalization. To mitigate noise fitting to irregular sampling, we introduce spectral-regularized Koopman dynamics modeling. It applies spectral-aware filtering in a latent space to extract denoised low-frequency trajectories and learns a compact Koopman operator to model system dynamics in a robust, linearized space. To handle non-stationarity in historical contexts, we design a context-informed heterogeneous pattern extraction mechanism. A target-conditioned attention module enables each future timestep to attend to distinct subsets of the historical windows, while an environmental signature derived from the current evolutionary pattern guides a learned router to adaptively identify the most predictive past context. 
Extensive experiments across eight classification and regression benchmarks show that AdaSpecK achieves SOTA performance.

\subsection*{AI use statement}


In this work, we used generative AI tools to improve the clarity, grammar, and overall quality of the writing.
We have not used generative AI tools for generating the core ideas, formulating the research questions, and developing the methodology. We have reviewed and edited all AI-assisted work. In particular, we examined the correctness, logical coherence, and contextual appropriateness of any AI-generated text to ensure that it accurately reflected our intended meaning and met academic standards. We take responsibility for the final content of this work,
including text, claims or artifacts produced with the aid of generative AI.



\subsection*{Ethics statement}
This work is conducted entirely on publicly available benchmark datasets, as detailed in the paper, and does not involve the release of any personal or sensitive information. 
No human subjects are involved in this research, ensuring that our work complies with ethical standards in research integrity.

\subsection*{Reproducibility statement}
All experiments in this work are conducted on publicly available benchmark datasets, and the reported performance of AdaSpecK is fully reproducible. Details of model architecture, training procedures, and evaluation protocols are provided in the main text and appendix. 
To further facilitate reproducibility, we release the code and datasets at \href{}{https://anonymous.4open.science/r/Ada-Spec-K}.

\bibliography{iclr2027_conference}

@inproceedings{DBLP:conf/icml/dg1,
  author       = {Krikamol Muandet and
                  David Balduzzi and
                  Bernhard Sch{\"{o}}lkopf},
  title        = {Domain Generalization via Invariant Feature Representation},
  booktitle    = {{ICML}},
  pages        = {10--18},
  year         = {2013},
}

@inproceedings{DBLP:conf/iccv/dg2,
  author       = {Saeid Motiian and
                  Marco Piccirilli and
                  Donald A. Adjeroh and
                  Gianfranco Doretto},
  title        = {Unified Deep Supervised Domain Adaptation and Generalization},
  booktitle    = {{ICCV}},
  pages        = {5716--5726},
  year         = {2017},
}

@inproceedings{DBLP:conf/iccv/dg3,
  author       = {Da Li and
                  Yongxin Yang and
                  Yi{-}Zhe Song and
                  Timothy M. Hospedales},
  title        = {Deeper, Broader and Artier Domain Generalization},
  booktitle    = {{ICCV}},
  pages        = {5543--5551},
  year         = {2017},
}

@inproceedings{DBLP:conf/nips/dg4,
  author       = {Yogesh Balaji and
                  Swami Sankaranarayanan and
                  Rama Chellappa},
  title        = {MetaReg: Towards Domain Generalization using Meta-Regularization},
  booktitle    = {NeurIPS},
  pages        = {1006--1016},
  year         = {2018},
}

@inproceedings{DBLP:conf/nips/dg5,
  author       = {Qi Dou and
                  Daniel Coelho de Castro and
                  Konstantinos Kamnitsas and
                  Ben Glocker},
  title        = {Domain Generalization via Model-Agnostic Learning of Semantic Features},
  booktitle    = {NeurIPS},
  pages        = {6447--6458},
  year         = {2019},
}

@inproceedings{DBLP:conf/icdm/dg6,
  author       = {Dazhou Yu and
                  Guangji Bai and
                  Yun Li and
                  Liang Zhao},
  editor       = {Xingquan Zhu and
                  Sanjay Ranka and
                  My T. Thai and
                  Takashi Washio and
                  Xindong Wu},
  title        = {Deep Spatial Domain Generalization},
  booktitle    = {{ICDM}},
  pages        = {1293--1298},
  publisher    = {{IEEE}},
  year         = {2022},
}

@inproceedings{DBLP:conf/iclr/dg7,
  author       = {Ishaan Gulrajani and
                  David Lopez{-}Paz},
  title        = {In Search of Lost Domain Generalization},
  booktitle    = {{ICLR}},
  year         = {2021},
}

@inproceedings{DBLP:conf/nips/manip1,
  author       = {Alexander H. Liu and
                  Yen{-}Cheng Liu and
                  Yu{-}Ying Yeh and
                  Yu{-}Chiang Frank Wang},
  title        = {A Unified Feature Disentangler for Multi-Domain Image Translation
                  and Manipulation},
  booktitle    = {NeurIPS},
  pages        = {2595--2604},
  year         = {2018},
}

@inproceedings{DBLP:conf/cvpr/manip2,
  author       = {Fengchun Qiao and
                  Long Zhao and
                  Xi Peng},
  title        = {Learning to Learn Single Domain Generalization},
  booktitle    = {{CVPR}},
  pages        = {12553--12562},
  publisher    = {Computer Vision Foundation / {IEEE}},
  year         = {2020},
}

@inproceedings{DBLP:conf/cvpr/rep1,
  author       = {Rui Gong and
                  Wen Li and
                  Yuhua Chen and
                  Luc Van Gool},
  title        = {{DLOW:} Domain Flow for Adaptation and Generalization},
  booktitle    = {{CVPR}},
  pages        = {2477--2486},
  publisher    = {Computer Vision Foundation / {IEEE}},
  year         = {2019},
}

@article{DBLP:journals/pami/rep2,
  author       = {Wen Li and
                  Zheng Xu and
                  Dong Xu and
                  Dengxin Dai and
                  Luc Van Gool},
  title        = {Domain Generalization and Adaptation Using Low Rank Exemplar SVMs},
  journal      = {{IEEE} transactions on pattern analysis and machine intelligence},
  volume       = {40},
  number       = {5},
  pages        = {1114--1127},
  year         = {2018},
}

@inproceedings{DBLP:conf/cikm/meta1,
  author       = {Zekun Cai and
                  Renhe Jiang and
                  Xinyu Yang and
                  Zhaonan Wang and
                  Diansheng Guo and
                  Hill Hiroki Kobayashi and
                  Xuan Song and
                  Ryosuke Shibasaki},
  title        = {MemDA: Forecasting Urban Time Series with Memory-based Drift Adaptation},
  booktitle    = {{CIKM}},
  pages        = {193--202},
  year         = {2023},
}

@inproceedings{DBLP:conf/aaai/meta2,
  author       = {Da Li and
                  Yongxin Yang and
                  Yi{-}Zhe Song and
                  Timothy M. Hospedales},
  title        = {Learning to Generalize: Meta-Learning for Domain Generalization},
  booktitle    = {{AAAI}},
  pages        = {3490--3497},
  year         = {2018},
}

@inproceedings{DBLP:conf/icip/ensem1,
  author       = {Massimiliano Mancini and
                  Samuel Rota Bul{\`{o}} and
                  Barbara Caputo and
                  Elisa Ricci},
  title        = {Best Sources Forward: Domain Generalization through Source-Specific
                  Nets},
  booktitle    = {{ICIP}},
  pages        = {1353--1357},
  publisher    = {{IEEE}},
  year         = {2018},
}

@inproceedings{DBLP:conf/icml/disshift,
  author       = {Elise Han and
                  Chengpiao Huang and
                  Kaizheng Wang},
  title        = {Model Assessment and Selection under Temporal Distribution Shift},
  booktitle    = {{ICML}},
  series       = {Proceedings of Machine Learning Research},
  pages        = {17374--17392},
  year         = {2024},
}

@inproceedings{DBLP:conf/iclr/nonstat,
  author       = {Rui Wang and
                  Yihe Dong and
                  Sercan {\"{O}}. Arik and
                  Rose Yu},
  title        = {Koopman Neural Operator Forecaster for Time-series with Temporal Distributional
                  Shifts},
  booktitle    = {{ICLR}},
  year         = {2023},
}

@article{DBLP:journals/tkde/condrift1,
  author       = {Jie Lu and
                  Anjin Liu and
                  Fan Dong and
                  Feng Gu and
                  Jo{\~{a}}o Gama and
                  Guangquan Zhang},
  title        = {Learning under Concept Drift: {A} Review},
  journal      = {{IEEE} transactions on knowledge and data engineering},
  volume       = {31},
  number       = {12},
  pages        = {2346--2363},
  year         = {2019},
}

@inproceedings{DBLP:conf/icml/condrift2,
  author       = {Jingyuan Wang and
                  Zhimei Ren and
                  Ruohan Zhan and
                  Zhengyuan Zhou},
  title        = {Distributionally Robust Policy Learning under Concept Drifts},
  booktitle    = {{ICML}},
  series       = {PMLR},
  year         = {2025},
}

@article{DBLP:journals/tkde/dgcategory,
  author       = {Jindong Wang and
                  Cuiling Lan and
                  Chang Liu and
                  Yidong Ouyang and
                  Tao Qin and
                  Wang Lu and
                  Yiqiang Chen and
                  Wenjun Zeng and
                  Philip S. Yu},
  title        = {Generalizing to Unseen Domains: {A} Survey on Domain Generalization},
  journal      = {{IEEE} transactions on knowledge and data engineering},
  volume       = {35},
  number       = {8},
  pages        = {8052--8072},
  year         = {2023},
}

@inproceedings{DBLP:conf/eccv/smldg,
  author       = {Da Li and
                  Yongxin Yang and
                  Yi{-}Zhe Song and
                  Timothy M. Hospedales},
  editor       = {Adrien Bartoli and
                  Andrea Fusiello},
  title        = {Sequential Learning for Domain Generalization},
  booktitle    = {{ECCV} 2020 Workshops},
  series       = {Lecture Notes in Computer Science},
  pages        = {603--619},
  publisher    = {Springer},
  year         = {2020},
}

@article{zeng2024koopmantknets,
  author       = {Qiuhao Zeng and Wei Wang and Fan Zhou and Gezheng Xu and Ruizhi Pu and Changjian Shui and Christian Gagn{\'e} and Shichun Yang and Charles X. Ling and Boyu Wang},
  title        = {Generalizing across Temporal Domains with Koopman Operators},
  journal      = {AAAI},
  volume       = {38},
  number       = {15},
  pages        = {16651--16659},
  year         = {2024},
}

@inproceedings{DBLP:conf/nips/GradientI,
  author       = {Anshul Nasery and
                  Soumyadeep Thakur and
                  Vihari Piratla and
                  Abir De and
                  Sunita Sarawagi},
  title        = {Training for the Future: {A} Simple Gradient Interpolation Loss to
                  Generalize Along Time},
  booktitle    = {NeurIPS},
  pages        = {19198--19209},
  year         = {2021},
}

@inproceedings{DBLP:conf/icml/lssae,
  author       = {Tiexin Qin and
                  Shiqi Wang and
                  Haoliang Li},
  title        = {Generalizing to Evolving Domains with Latent Structure-Aware Sequential
                  Autoencoder},
  booktitle    = {{ICML}},
  series       = {Proceedings of Machine Learning Research},
  pages        = {18062--18082},
  publisher    = {{PMLR}},
  year         = {2022},
}

@inproceedings{bai2023drain,
  author       = {Guangji Bai and Chen Ling and Liang Zhao},
  title        = {Temporal Domain Generalization with Drift-Aware Dynamic Neural Networks},
  booktitle    = {ICLR},
  year         = {2023},
}

@inproceedings{DBLP:conf/nips/CaiBJ0024koodo,
  author       = {Zekun Cai and
                  Guangji Bai and
                  Renhe Jiang and
                  Xuan Song and
                  Liang Zhao},
  title        = {Continuous Temporal Domain Generalization},
  booktitle    = {NeurIPS},
  year         = {2024},
}

@article{yu2025frekoo,
  author       = {En Yu and Jie Lu and Xiaoyu Yang and Guangquan Zhang and Zhen Fang},
  title        = {Learning Robust Spectral Dynamics for Temporal Domain Generalization},
  journal      = {arXiv preprint arXiv:2505.12585},
  year         = {2025},
}

@inproceedings{DBLP:conf/nips/Fan,
  author       = {Weiwei Ye and
                  Songgaojun Deng and
                  Qiaosha Zou and
                  Ning Gui},
  title        = {Frequency Adaptive Normalization For Non-stationary Time Series Forecasting},
  booktitle    = {NeurIPS},
  year         = {2024},
}

@article{DBLP:journals/corr/abs-1907-02893,
  author       = {Mart{\'{\i}}n Arjovsky and
                  L{\'{e}}on Bottou and
                  Ishaan Gulrajani and
                  David Lopez{-}Paz},
  title        = {Invariant Risk Minimization},
  journal      = {CoRR},
  volume       = {abs/1907.02893},
  year         = {2019},
}

@inproceedings{DBLP:conf/icml/KruegerCJ0BZPC21,
  author       = {David Krueger and
                  Ethan Caballero and
                  J{\"{o}}rn{-}Henrik Jacobsen and
                  Amy Zhang and
                  Jonathan Binas and
                  Dinghuai Zhang and
                  R{\'{e}}mi Le Priol and
                  Aaron C. Courville},
  title        = {Out-of-Distribution Generalization via Risk Extrapolation ({REx})},
  booktitle    = {{ICML}},
  pages        = {5815--5826},
  publisher    = {{PMLR}},
  year         = {2021},
}

@article{DBLP:journals/csur/GamaZBPB14,
  author       = {Jo{\~{a}}o Gama and
                  Indre Zliobaite and
                  Albert Bifet and
                  Mykola Pechenizkiy and
                  Abdelhamid Bouchachia},
  title        = {A survey on concept drift adaptation},
  journal      = {{ACM} Comput. Surv.},
  volume       = {46},
  number       = {4},
  pages        = {44:1--44:37},
  year         = {2014},
}

@inproceedings{DBLP:conf/nips/ChenRBD18,
  author       = {Tian Qi Chen and
                  Yulia Rubanova and
                  Jesse Bettencourt and
                  David Duvenaud},
  title        = {Neural Ordinary Differential Equations},
  booktitle    = {NeurIPS 2018},
  pages        = {6572--6583},
  year         = {2018},
}

@article{DBLP:journals/corr/abs-1907-03907,
  author       = {Yulia Rubanova and
                  Ricky T. Q. Chen and
                  David Duvenaud},
  title        = {Latent ODEs for Irregularly-Sampled Time Series},
  journal      = {CoRR},
  volume       = {abs/1907.03907},
  year         = {2019},
  url          = {http://arxiv.org/abs/1907.03907},
  eprinttype   = {arXiv},
  eprint       = {1907.03907},
}

@inproceedings{DBLP:conf/nips/Autoformer,
  author       = {Haixu Wu and
                  Jiehui Xu and
                  Jianmin Wang and
                  Mingsheng Long},
  title        = {Autoformer: Decomposition Transformers with Auto-Correlation for Long-Term
                  Series Forecasting},
  booktitle    = {NeurIPS},
  pages        = {22419--22430},
  year         = {2021},
}

@inproceedings{DBLP:conf/icml/FEDformer,
  author       = {Tian Zhou and
                  Ziqing Ma and
                  Qingsong Wen and
                  Xue Wang and
                  Liang Sun and
                  Rong Jin},
  title        = {FEDformer: Frequency Enhanced Decomposed Transformer for Long-term
                  Series Forecasting},
  booktitle    = {{ICML}},
  pages        = {27268--27286},
  publisher    = {{PMLR}},
  year         = {2022},
}

@inproceedings{DBLP:conf/nips/fiternet,
  author       = {Kun Yi and
                  Jingru Fei and
                  Qi Zhang and
                  Hui He and
                  Shufeng Hao and
                  Defu Lian and
                  Wei Fan},
  title        = {FilterNet: Harnessing Frequency Filters for Time Series Forecasting},
  booktitle    = {NeurIPS},
  year         = {2024},
}

@inproceedings{DBLP:conf/iclr/timesnet,
  author       = {Haixu Wu and
                  Tengge Hu and
                  Yong Liu and
                  Hang Zhou and
                  Jianmin Wang and
                  Mingsheng Long},
  title        = {TimesNet: Temporal 2D-Variation Modeling for General Time Series Analysis},
  booktitle    = {{ICLR}},
  year         = {2023},
}

@article{koopman1931,
  author       = {B. O. Koopman},
  title        = {Hamiltonian Systems and Transformation in Hilbert Space},
  journal      = {Proceedings of the National Academy of Sciences},
  volume       = {17},
  number       = {5},
  pages        = {315--318},
  year         = {1931},
  doi          = {10.1073/pnas.17.5.315},
}

@article{DBLP:journals/siamrev/koopman2022,
  author       = {Steven L. Brunton and
                  Marko Budisic and
                  Eurika Kaiser and
                  J. Nathan Kutz},
  title        = {Modern Koopman Theory for Dynamical Systems},
  journal      = {{SIAM} Rev.},
  volume       = {64},
  number       = {2},
  pages        = {229--340},
  year         = {2022},
}

@inproceedings{pmlr-v119-wang20h,
  title        = {Continuously Indexed Domain Adaptation},
  author       = {Wang, Hao and He, Hao and Katabi, Dina},
  booktitle    = {ICML},
  pages        = {9898--9907},
  year         = {2020},
  volume       = {119},
  publisher    = {PMLR},
}

@article{DBLP:journals/spm/mnist,
  author       = {Li Deng},
  title        = {The {MNIST} Database of Handwritten Digit Images for Machine Learning
                  Research [Best of the Web]},
  journal      = {{IEEE} Signal Process. Mag.},
  volume       = {29},
  number       = {6},
  pages        = {141--142},
  year         = {2012},
}

@article{DBLP:journals/tkde/twitter1,
  author       = {Liang Zhao and
                  Qian Sun and
                  Jieping Ye and
                  Feng Chen and
                  Chang{-}Tien Lu and
                  Naren Ramakrishnan},
  title        = {Feature Constrained Multi-Task Learning Models for Spatiotemporal
                  Event Forecasting},
  journal      = {{IEEE} Transactions on Knowledge and Data Engineering},
  volume       = {29},
  number       = {5},
  pages        = {1059--1072},
  year         = {2017},
}

@inproceedings{DBLP:conf/sdm/twitter2,
  author       = {Guangji Bai and
                  Johnny Torres and
                  Junxiang Wang and
                  Liang Zhao and
                  Cristina L. Abad and
                  Carmen Vaca},
  title        = {Sign-Regularized Multi-Task Learning},
  booktitle    = {Proceedings of the 2023 {SIAM} International Conference on Data Mining, {SDM} },
  pages        = {793--801},
  publisher    = {{SIAM}},
  year         = {2023},
}

@inproceedings{yao2022wildtime,
  author       = {Huaxiu Yao and Caroline Choi and Bochuan Cao and Yoonho Lee and Pang Wei Koh and Chelsea Finn},
  title        = {Wild-Time: A Benchmark of in-the-Wild Distribution Shift over Time},
  booktitle    = {NeurIPS},
  volume       = {35},
  pages        = {10309--10324},
  year         = {2022},
}

@inproceedings{DBLP:conf/kdd/Cyclone,
  author       = {Boyo Chen and
                  Buo{-}Fu Chen and
                  Hsuan{-}Tien Lin},
  title        = {Rotation-blended CNNs on a New Open Dataset for Tropical Cyclone Image-to-intensity
                  Regression},
  booktitle    = {{SIGKDD}},
  pages        = {90--99},
  publisher    = {{ACM}},
  year         = {2018},
}

@incollection{fernandes2015proactive,
  author       = {Kelwin Fernandes and Pedro Vinagre and Paulo Cortez},
  title        = {A Proactive Intelligent Decision Support System for Predicting the Popularity of Online News},
  booktitle    = {Progress in Artificial Intelligence},
  series       = {Lecture Notes in Computer Science},
  volume       = {9273},
  pages        = {535--546},
  publisher    = {Springer},
  year         = {2015},
  doi          = {10.1007/978-3-319-23485-4_53},
}

@article{candanedo2017data,
  author       = {Luis M. Candanedo and V{\'e}ronique Feldheim and Dominique Deramaix},
  title        = {Data Driven Prediction Models of Energy Use of Appliances in a Low-Energy House},
  journal      = {Energy and Buildings},
  volume       = {140},
  pages        = {81--97},
  year         = {2017},
  doi          = {10.1016/j.enbuild.2017.01.083},
}

@article{decock2011ames,
  author       = {Dean De Cock},
  title        = {Ames, Iowa: Alternative to the Boston Housing Data as an End of Semester Regression Project},
  journal      = {Journal of Statistics Education},
  volume       = {19},
  number       = {3},
  year         = {2011},
  doi          = {10.1080/10691898.2011.11889627},
}

@misc{statlog_shuttle_uci,
  title        = {Statlog (Shuttle) [Dataset]},
  author       = {{UCI Machine Learning Repository}},
  year         = {2025},
  doi          = {10.24432/C5WS31},
  url          = {https://archive.ics.uci.edu/dataset/148/statlog+shuttle},
  note         = {Accessed: 2026-05-02}
}
\bibliographystyle{iclr2027_conference}

\clearpage
\appendix
\section{Appendix}
\subsection{Datasets}
\label{app:data}
Following Koodos~\citep{DBLP:conf/nips/CaiBJ0024koodo} and FreKoo~\citep{yu2025frekoo}, and focusing on the datasets that are both discussed in the two reference papers and implemented in the new codebase, we evaluate AdaSpecK on eight benchmarks covering classification and regression under different temporal drift patterns. Their main characteristics are summarized as follows:

\begin{itemize}
  \item \textbf{Rotated MNIST (Rot-MNIST).} This is a synthetic image classification benchmark built from MNIST digits~\citep{DBLP:journals/spm/mnist}, where concept drift is created by gradually rotating the images across domains as in the temporal benchmark protocols of Koodos~\citep{DBLP:conf/nips/CaiBJ0024koodo} and FreKoo~\citep{yu2025frekoo}. The domains are temporally ordered, and the prediction target always lies in later unseen domains. Its main difficulty comes from a steadily increasing geometric transformation rather than label noise.
  \item \textbf{Twitter.} This dataset uses tweet streams to estimate influenza activity. The domains correspond to temporally separated collection windows during multiple flu seasons, so the drift is driven by changes in tweet volume, seasonal disease activity, and evolving user behavior~\citep{DBLP:conf/nips/CaiBJ0024koodo,yu2025frekoo}. Compared with synthetic rotation datasets, it reflects a real-world temporally evolving classification problem with stronger non-stationarity.
  \item \textbf{YearBook.} This is a gender classification benchmark built from frontal yearbook portraits collected over many decades, following the YearBook temporal-shift benchmark in Wild-Time~\citep{yao2022wildtime}. Each domain corresponds to one sampled year in chronological order. The drift arises from long-term changes in clothing, hairstyle, imaging quality, and broader social norms, making it a representative visual benchmark with slow but persistent temporal shift.
  \item \textbf{Cyclone.} This regression dataset predicts tropical cyclone intensity from satellite imagery. Each domain is associated with temporally ordered cyclone observations, and the shift is induced by seasonal and regional variations in storm appearance and wind intensity ~\citep{DBLP:conf/nips/CaiBJ0024koodo,yu2025frekoo}. It is therefore a visually complex forecasting benchmark with substantial temporal variation.
  \item \textbf{House.} This regression benchmark predicts house prices from tabular features collected over multiple years. The domains are organized chronologically, and the underlying drift mainly reflects changing market conditions and temporal economic variation, making it a real-world nonstationary tabular dataset ~\citep{decock2011ames,DBLP:conf/nips/CaiBJ0024koodo,yu2025frekoo}.
  \item \textbf{Online News Popularity (ONP).} This classification dataset contains news articles from Mashable with 58 heterogeneous features, and the task is to predict article popularity~\citep{fernandes2015proactive}. The data are split into time-ordered domains over roughly two years, with the first five domains used for training and the last one for testing in the TDG protocol. Its drift is relatively mild compared with the other benchmarks, but it still captures gradual changes in how content attracts social attention over time.
  \item \textbf{Shuttle.} This dataset is a temporally ordered multi-class classification benchmark derived from flight-status data and is known for its strong class imbalance~\citep{statlog_shuttle_uci,DBLP:conf/nips/CaiBJ0024koodo,yu2025frekoo}. The domains are formed chronologically, with earlier periods used for training and the latest period used for testing. The challenge comes from real-world distribution changes in operational patterns over time rather than from manually constructed perturbations.
  \item \textbf{ApplianceEnergy (Appliance).} This regression benchmark predicts appliance energy consumption from indoor sensor readings in a low-energy building~\citep{candanedo2017data}. The data are recorded at 10-minute intervals over several months and then partitioned into chronological domains. The drift is associated with changing usage behavior and environmental conditions across time, which makes it a practical benchmark for periodic and real-world temporal variation.
\end{itemize}

\subsection{Baseline}

We compare AdaSpecK with several families of baselines that reflect different ways of handling temporal drift. Offline, LastDomain, and IncFinetune follow the standard non-temporal or sequential training baselines used in recent TDG studies \citep{DBLP:conf/nips/CaiBJ0024koodo,yu2025frekoo}, where Offline trains a single model on the union of all source domains and ignores temporal order entirely, LastDomain uses only the most recent observed source domain to test whether short-range recency alone is sufficient, and IncFinetune updates the predictor sequentially from earlier to later domains as a practical continual-training baseline. IRM and V-REx are invariant-learning methods that aim to suppress domain-specific shortcuts by favoring predictors whose risks are stable across source domains \citep{DBLP:journals/corr/abs-1907-02893,DBLP:conf/icml/KruegerCJ0BZPC21}, although they do not explicitly model domain evolution over time. CIDA incorporates temporal or domain information as a continuous conditioning variable and aligns representations across domains \citep{pmlr-v119-wang20h}, while TKNets models transitions between temporally ordered domains through a dedicated temporal transfer mechanism \citep{zeng2024koopmantknets}. DRAIN and DRAIN-$\Delta t$ forecast future predictors by learning parameter dynamics from the observed domain sequence, with the latter additionally incorporating time-gap information for non-uniform intervals \citep{bai2023drain}. DeepODE models temporal evolution in a latent dynamical system defined by neural ordinary differential equations \citep{DBLP:conf/nips/ChenRBD18,DBLP:journals/corr/abs-1907-03907}, and Koodos introduces Koopman-space parameter evolution with structured dynamic constraints \citep{DBLP:conf/nips/CaiBJ0024koodo}. We also report FreKoo as a strong recent temporal generalization reference that combines spectral decomposition and Koopman modeling to separate low-frequency temporal structure from high-frequency variations \citep{yu2025frekoo}. Together, these methods cover time-agnostic training, invariant learning, temporal alignment, parameter-forecasting approaches, and frequency-aware temporal dynamics.

\subsection{Implementation Details}
\label{app:impl}
All results were averaged over 5 independent runs and we report the mean and standard deviation. We implement AdaSpecK in PyTorch and conduct all experiments on a 64-bit machine with an NVIDIA GeForce RTX 3090 GPU. Consistent with the summary in the main text, we use Adam with cosine learning-rate annealing, set the environment embedding dimension to 32, fix the low-frequency ratio to 0.40, use RK4 with step size 0.2, adopt a learning rate of 0.0005 and weight decay of 0.005 across datasets, and use 4 experts for Twitter and the relatively simpler tabular benchmarks (House, ONP, and Appliance) and 8 experts for Rotated MNIST, YearBook, Cyclone, and Shuttle. The remaining dataset-specific architectural and dimensional settings are as follows:
\begin{enumerate}
    \item \textbf{Rotated MNIST.} We use 35 training domains, temporal segments of length 20, and a 15-step target horizon. The shared backbone is a 3-layer CNN with channel widths $[32,32,64]$, where each convolution is followed by ReLU and max pooling, and a dropout layer is applied after flattening. We train for 200 epochs with batch size 1. The latent dimension is 64, the hidden dimension is 256, the number of experts is 8, the spectral bin number is 5, and the Koopman rollout length is 15.
    \item \textbf{Twitter.} We use 35 training domains, segment length 20, and a 15-step target horizon. The backbone first embeds the categorical state feature with a 10-dimensional embedding and then applies a 3-layer MLP with hidden widths $256$ and $128$ to produce a 64-dimensional representation. We train for 200 epochs with batch size 1. The latent dimension is 64, the hidden dimension is 128, the number of experts is 4, the spectral bin number is 4, and the Koopman rollout length is 10.
    \item \textbf{YearBook.} We use 28 training domains, segment length 20, and a 12-step target horizon. The backbone is a 3-stage CNN with batch normalization and channel widths $[64,128,256]$, followed by a linear projection to 1024 dimensions and dropout. We train for 170 epochs with batch size 1. The latent dimension is 128, the hidden dimension is 256, the number of experts is 8, the spectral bin number is 5, and the Koopman rollout length is 15.
    \item \textbf{Cyclone.} We use 50 training domains, segment length 20, and a 22-step target horizon. The backbone is a 4-layer CNN with channel widths $[32,32,64,64]$, where each block uses ReLU and max pooling, followed by flattening and dropout. We train for 100 epochs with batch size 1. The latent dimension is 128, the hidden dimension is 384, the number of experts is 8, the spectral bin number is 6, and the Koopman rollout length is 15.
    \item \textbf{House.} We use 28 training domains, segment length 20, and a 12-step target horizon. For this tabular dataset, the backbone is the identity mapping and the predictor operates directly on the 50-dimensional input features. We train for 100 epochs with batch size 1. The latent dimension is 64, the hidden dimension is 256, the number of experts is 4, the spectral bin number is 5, and the Koopman rollout length is 6.
    \item \textbf{ONP.} We use 5 training domains, temporal segments of length 4, and a 1-step target horizon. The model uses the raw 58-dimensional tabular features as input without an additional backbone. We train for 700 epochs with batch size 2. The latent dimension is 64, the hidden dimension is 256, the number of experts is 4, the spectral bin number is 5, and the Koopman rollout length is 3.
    \item \textbf{Shuttle.} We use 7 training domains, temporal segments of length 4, and a 1-step target horizon. The model takes the 9-dimensional tabular features directly as input. We train for 350 epochs with batch size 2. The latent dimension is 64, the hidden dimension is 256, the number of experts is 8, the spectral bin number is 5, and the Koopman rollout length is 5.
    \item \textbf{Appliance.} We use 8 training domains, temporal segments of length 4, and a 1-step target horizon. The model directly uses the 37-dimensional tabular features. We train for 500 epochs with batch size 2. The latent dimension is 64, the hidden dimension is 256, the number of experts is 4, the spectral bin number is 5, and the Koopman rollout length is 3.
\end{enumerate}

\end{document}